\documentclass[journal, letterpaper]{IEEEtran}
\usepackage{graphicx}
\usepackage{url}
\usepackage{cite}
\usepackage{amsmath}
\usepackage{amssymb}
\usepackage{textgreek}
\usepackage{listings}
\usepackage{csvsimple}
\usepackage{longtable}
\usepackage{color} 
\usepackage{booktabs}

\title{Quantized SO(3)-Equivariant Graph Neural Networks for Efficient Molecular Property Prediction}

\author{Haoyu Zhou, Ping Xue, Hao Zhang, and Tianfan Fu%
\thanks{ P. Xue, H. Zhang are with Suzhou Lab, Suzhou, China.}%
\thanks{H. Zhou,T. Fu is with Nanjing University, Nanjing, China.}%
}
\begin{document}
\markboth{Course Name/Number: Project Name}{}
\maketitle

\begin{figure*}[t]
  \centering
  \includegraphics[width=\linewidth]{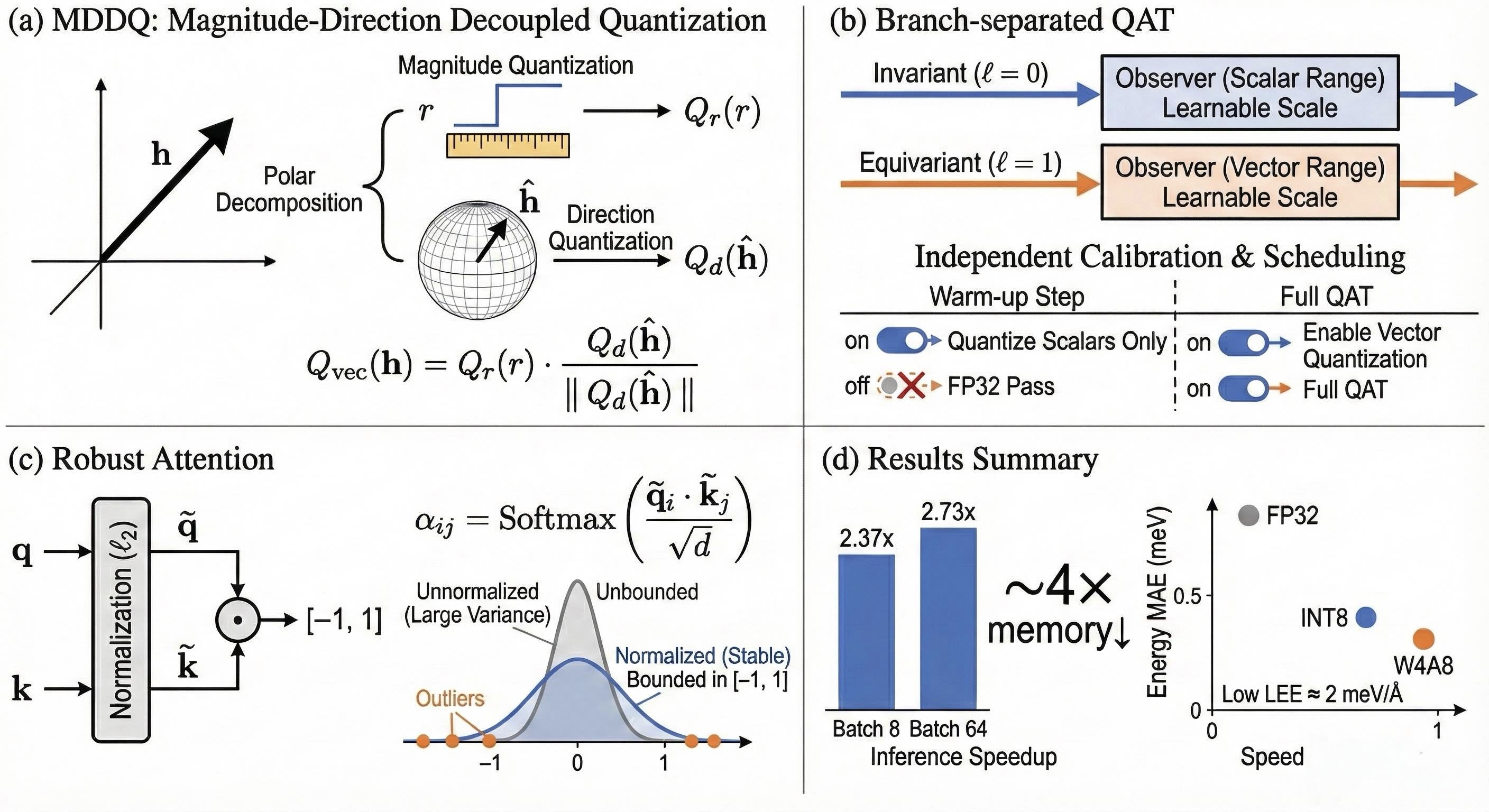}
  \caption{\textbf{Overview of the proposed equivariant quantization framework.} 
  (a) \textbf{MDDQ}: Decouples equivariant vectors into magnitude $r$ and direction $\hat{\mathbf{h}}$ to preserve geometric orientation under low precision. 
  (b) \textbf{Branch-separated QAT}: Treats invariant and equivariant features differently with a staged training schedule. 
  (c) \textbf{Robust Attention}: Stabilizes dot-products via $\ell_2$ normalization of queries and keys. 
  (d) \textbf{Results Summary}: Our method achieves $2.37$--$2.73\times$ faster inference and $\sim 4\times$ memory reduction with accuracy comparable to FP32 models.}
  \label{fig:teaser}
\end{figure*}

\begin{abstract}
Deploying 3D graph neural networks (GNNs) that are equivariant to 3D rotations (the group $SO(3)$) on edge devices is challenging due to their high computational cost. This paper addresses the problem by compressing and accelerating an $SO(3)$-equivariant GNN using low-bit quantization techniques. Specifically, we introduce three innovations for quantized equivariant transformers: (1) a \textbf{magnitude-direction decoupled quantization} scheme that separately quantizes the norm and orientation of equivariant (vector) features, (2) a \textbf{branch-separated quantization-aware training} strategy that treats invariant and equivariant feature channels differently in an attention-based $SO(3)$-GNN, and (3) a robustness-enhancing \textbf{attention normalization} mechanism that stabilizes low-precision attention computations. Experiments on the QM9 and rMD17 molecular benchmarks demonstrate that our 8-bit models achieve accuracy on energy and force predictions comparable to full-precision baselines with markedly improved efficiency. We also conduct ablation studies to quantify the contribution of each component to maintain accuracy and equivariance under quantization, using the Local error of equivariance (LEE) metric. The proposed techniques enable the deployment of symmetry-aware GNNs in practical chemistry applications with 2.37--2.73$\times$ faster inference and $\sim$4$\times$ smaller model size, without sacrificing accuracy or physical symmetry.
\end{abstract}

\section{Introduction}

Equivariant GNNs respect the symmetries of 3D space and have driven rapid progress in molecular modeling and force-field learning~\cite{schutt2017schnet,thomas2018tensor,worrall2017harmonic,cohen2016group,weiler20183d,cohen2018spherical,cohen2019gauge,fuchs2020se,finzi2021practical,schutt2021equivariant,batzner20223,frank2022so3krates,frank2024euclidean,satorras2021n,unke2021spookynet}.
Yet, their heavy tensor algebra and attention operations strain edge hardware.
In parallel, low-bit quantization has become a standard efficiency tool for neural networks~\cite{zhou2016dorefa,choi2018pact,esser2019learned,jacob2018quantization,tailor2020degree,wei2022qdrop,xiao2023smoothquant,yao2022zeroquant}, but na\"ive application to equivariant models can degrade symmetry and accuracy.
Molecular property prediction from structure is a cornerstone of computational chemistry and materials science. Recent advancements in equivariant graph neural networks (GNNs) have enabled highly accurate predictions by respecting the underlying physical symmetries in molecular systems~\cite{batzner20223,frank2024euclidean}. In particular, networks equivariant to 3D rotations (the rotation group $SO(3)$) such as NequIP~\cite{batzner20223}, So3krates~\cite{frank2022so3krates,frank2024euclidean}, and $SE(3)$-Transformers~\cite{fuchs2020se} process 3D molecular graphs in a symmetry-aware manner, ensuring that rotating a molecule in space yields correspondingly rotated model predictions (\textit{e.g.}, for force vectors). These models achieve state-of-the-art accuracy on tasks like predicting quantum energies and forces~\cite{batzner20223,frank2024euclidean}. However, their complexity (\textit{e.g.}, tensor operations on spherical harmonics or global attention across many channels) makes them computationally heavy. This poses a barrier to deploying such models on resource-constrained devices (\textit{e.g.}, mobile phones or lab-on-chip sensors), which could enable \emph{in situ} chemistry analyses.

From a deployment viewpoint, three key challenges remain:\\ 
\noindent (1) \textbf{Computational burden} – $SO(3)$-equivariant representations (tensors, spherical-harmonic features, etc.) inflate FLOPs and memory usage; \\
\noindent (2) \textbf{Quantization brittleness} – naively applying standard 8-bit quantization or post-training quantization (PTQ) to equivariant GNNs can distort vector feature directions and magnitudes, causing significant accuracy drops and broken equivariance; \\
\noindent (3) \textbf{Uniform treatment of scalar/vector features} – treating all features identically during quantization is suboptimal, since invariant scalars and equivariant vectors have very different distributions and roles, requiring calibrated handling. This work targets these gaps with an equivariance-aware quantization framework that preserves symmetry and accuracy while enabling real-time low-bit inference;\\
\noindent (4) \textbf{Limitations of existing methods.} Generic quantization techniques developed for CNNs/Transformers~\cite{jacob2018quantization, zhou2016dorefa, wei2022qdrop} do not account for rotational structure; applied naively to equivariant GNNs, they induce orientation drift in vector features and unstable attention patterns. GNN-specific quantizers such as Degree-Quant~\cite{tailor2020degree} mitigate degree-induced scale variation, but they operate on scalar message-passing and do not preserve the geometry of $\ell>0$ (vector/tensor) representations, leaving equivariance fragile under low bit-widths. In short, no existing approach jointly addresses (a) preserving vector direction \emph{and} magnitude, (b) maintaining stable low-bit attention, and (c) differential treatment between invariant and equivariant branches.

To address these issues, 
in this paper, we develop a quantization scheme specifically tailored for $SO(3)$-equivariant GNNs (in particular, transformer-based architectures like So3krates~\cite{frank2022so3krates, frank2024euclidean}) and demonstrate its effectiveness. Our main contributions are summarized as follows:

\noindent (1) \textbf{Magnitude-Direction Decoupled Quantization (MDDQ).} We propose a novel quantization strategy that decouples the magnitude and direction components of vector (equivariant) features. By quantizing the norm and orientation separately, the rotational information in $SO(3)$-equivariant features is preserved even under low precision.

\noindent (2) \textbf{Branch-Separated QAT for Invariant/Equivariant Features.} We introduce a branch-specific quantization-aware training (QAT) strategy that treats invariant (scalar, $\ell=0$) and equivariant (vector, $\ell=1$) feature channels differently. Each branch uses dedicated fake quantization parameters and a staged training schedule, resulting in better calibrated quantization that respects each branch’s distribution and symmetry role.

\noindent (3) \textbf{Robust Attention Normalization.} We propose an $L^2$ normalization of attention queries and keys (analogous to cosine attention~\cite{henry2020query}) to stabilize the attention mechanism under quantization. This ensures that low-precision dot products are bounded and dominated by directional similarity rather than scale, greatly improving the robustness of attention computations in our INT8 models.

This is the first work to address quantization of $SO(3)$-equivariant graph neural networks. We show that an 8-bit equivariant transformer can achieve accuracy on par with its 32-bit counterpart on challenging molecular prediction tasks, while obtaining $4\times$ faster inference and $4\times$ reduction in model size in theory. The results point toward practical deployment of equivariant GNNs on low-power devices. For example, one can envision a mobile chemistry assistant that processes a molecule's structure and rapidly predicts its properties on-device. More broadly, our quantization approach provides a foundation for compressing other symmetry-preserving models without sacrificing their key physical guarantees. Finally, Fig.~\ref{fig:teaser} summarizes our core innovations: (a) MDDQ, (b) branch-separated QAT, (c) robust attention normalization, and (d) a summary of the resulting efficiency and accuracy improvements.

\begin{figure}[t]
  \centering
  \includegraphics[width=\linewidth]{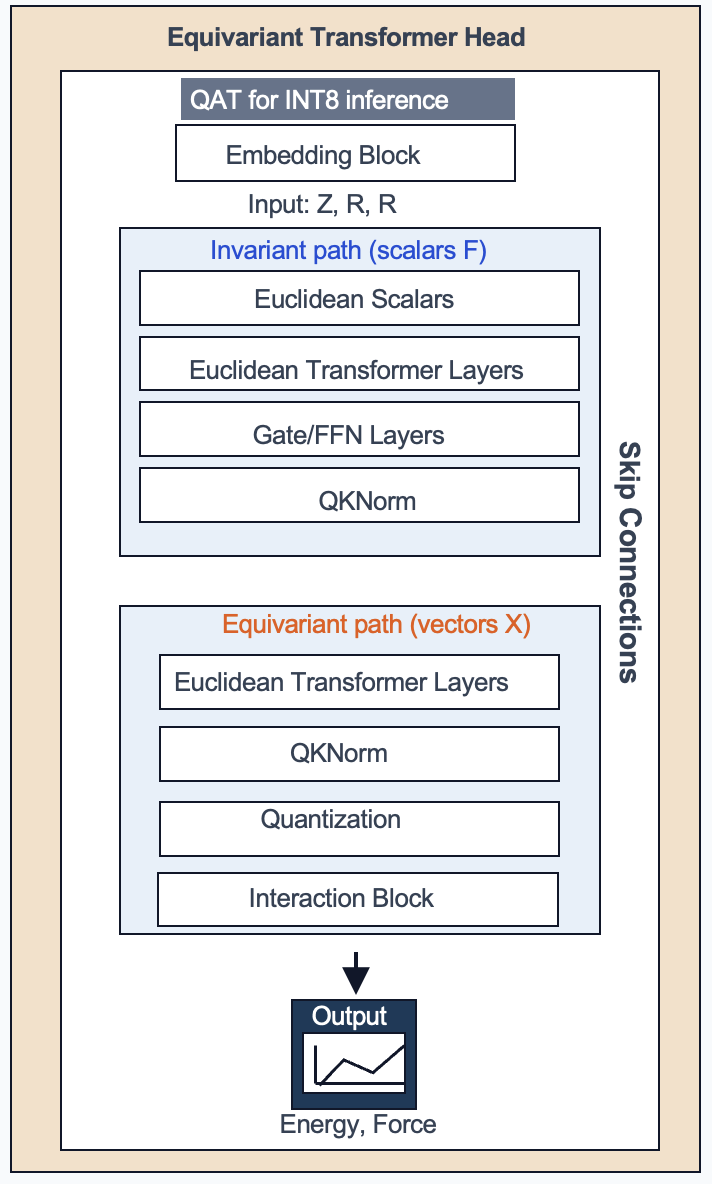}
  \caption{\textbf{Detailed architecture and quantization pipeline.} Inputs $Z$ and $\mathbf{r}$ are processed through decoupled invariant and equivariant paths.}
  \label{fig:arch_pipeline}
\end{figure}

\section{Related Work}

\noindent\textbf{Equivariant Graph Neural Networks.}
A rich line of work incorporates physical symmetries into GNNs for molecular modeling. Early neural potentials such as SchNet~\cite{schutt2017schnet} used continuous-filter convolutions to respect translational invariance and achieved accurate energy/force prediction on small molecules~\cite{chmiela2017machine}. Rotational equivariance was then introduced via tensor representations: Tensor Field Networks~\cite{thomas2018tensor} and $SE(3)$-Transformers~\cite{fuchs2020se} leverage features transforming under $SO(3)$ (often parameterized by spherical harmonics). Beyond these, $E(3)$/$SE(3)$-equivariant architectures including EGNN~\cite{satorras2021n}, NequIP~\cite{batzner20223}, and So3krates~\cite{frank2022so3krates,frank2024euclidean} deliver strong accuracy and data efficiency by coupling equivariant message passing with learned radial filters and self-attention. While these models validate the benefit of symmetry for accuracy, most prior deployments assume full precision on GPUs. Our work instead targets \emph{deployment efficiency} by quantizing rotation-equivariant representations while preserving geometric fidelity.

\noindent\textbf{Graph Neural Network Quantization.}
Model compression via quantization is attractive due to ubiquitous low-bit hardware. \emph{Post-training quantization} converts a trained model to low precision using calibration data, whereas \emph{quantization-aware training} (QAT) simulates quantization during training to retain accuracy~\cite{jacob2018quantization}. Generic low-bit techniques (\textit{e.g.}, DoReFa~\cite{zhou2016dorefa}, QDrop~\cite{wei2022qdrop}) have been effective in CNNs/Transformers, and GNN-specific adaptations account for structural irregularity—Degree-Quant introduces degree-dependent scaling to stabilize QAT on graphs~\cite{tailor2020degree}. However, prior work largely ignores \emph{equivariant} GNNs whose features are vector/tensor-valued and must preserve both magnitude and direction under $SO(3)$ actions. We extend GNN quantization to this setting with a magnitude–direction decoupling strategy and attention-side normalization tailored for equivariant transformers, enabling low-bit deployment while maintaining rotational consistency.

\section{Method}
We propose an equivariance-aware quantization scheme for a transformer-style $SO(3)$-equivariant GNN, aiming for fast low-bit inference without sacrificing accuracy or symmetry. Our method builds on a So3krates-like architecture with separate invariant (scalar) and equivariant (vector) branches; we use standard attention implementations and e3nn notation, as these do not affect the core quantization approach.

\subsection{Equivariant GNN Architecture Overview}
Our base model follows So3krates~\cite{frank2022so3krates}. The detailed internal architecture and the specific locations of the quantization observers are illustrated in Fig.~\ref{fig:arch_pipeline} (formerly Fig. 1). Given a 3D molecular graph $G=(V,E)$ (atoms $V$, edges $E$ defined by bonds or a distance cutoff), we construct a fully-connected graph among atoms within a cutoff radius. Each atom $i$ has coordinates $\mathbf{r}_i$ and initial features (e.g., atom type, charge). Interatomic geometry (relative displacement $\mathbf{r}_{ij}=\mathbf{r}_j - \mathbf{r}_i$) is encoded via spherical harmonics in the attention mechanism~\cite{fuchs2020se}, ensuring rotation-equivariant message passing.

In each layer of this $SO(3)$-equivariant transformer, every node carries both invariant scalar features and equivariant vector features. For example, each atom $i$ has scalar features $h^{(0)}_i$ (such as a learned embedding) and vector features $h^{(1)}_i$ (3D vectors for $\ell=1$ representations, e.g. forces or dipoles). (Higher-order $\ell>1$ features can be used, but in practice $\ell \le 1$ is common for efficiency~\cite{frank2024euclidean}.) The transformer updates these in two branches that only interact via attention: self-attention is applied mainly to the scalar channels, and the vector channels are updated by equivariant message functions modulated by those attention weights~\cite{fuchs2020se,frank2022so3krates}. This separation of invariant and equivariant branches is crucial for our quantization approach.

\subsection{Magnitude-Direction Decoupled Quantization}
A key idea of our work is to quantize equivariant features while preserving their direction. We decouple the quantization of magnitude and direction (Fig.~\ref{fig:teaser}a). In an $SO(3)$-equivariant network, some features are vector-valued and rotate with the input. Consider a 3D feature vector $\mathbf{h}_i$ associated with atom $i$. Under a rotation $R \in SO(3)$ of the input, this feature should transform as $R\,\mathbf{h}_i$. We can decompose $\mathbf{h}_i$ into its magnitude $r_i = \|\mathbf{h}_i\|$ (a rotation-invariant scalar) and direction $\hat{\mathbf{h}}_i = \mathbf{h}_i/\|\mathbf{h}_i\|$. We separate each vector into its magnitude $r_i$ and direction $\hat{\mathbf{h}}_i$ to ensure directional information is preserved. Standard per-component quantization (e.g., rounding each component of $\mathbf{h}_i$ to 8-bit) can disrupt the vector's overall magnitude and orientation. For example, a small vector could collapse to zero if all three components fall below the quantization step size.

We quantize $r_i$ with a $b$-bit scalar quantizer $Q_r(\cdot)$ (with a learned step size $\Delta_r$ during QAT). Then we quantize the unit direction $\hat{\mathbf{h}}_i$ using a per-component quantizer $Q_d(\cdot)$, and re-normalize it to unit length. Specifically:
\begin{equation}
Q_{\text{vec}}(\mathbf{h}_i) = Q_r(r_i)\; \frac{Q_d(\hat{\mathbf{h}}_i)}{\|\,Q_d(\hat{\mathbf{h}}_i)\|\;}~,
\end{equation}
where $Q_r(r_i)$ is an 8-bit quantized magnitude and $Q_d(\hat{\mathbf{h}}_i)$ produces a 3D quantized direction vector. The reconstructed vector $Q_{\text{vec}}(\mathbf{h}_i)$ has an 8-bit magnitude and a direction constrained to the unit sphere. This decoupling preserves directional information: rotations of the input mainly affect $\hat{\mathbf{h}}_i$ (quantized uniformly over directions), while $r_i$ is quantized independently. Thus, small changes in orientation do not cause large jumps in the quantized magnitude (and vice versa). Geometrically, this is analogous to quantizing in spherical coordinates (angles and radius) instead of Cartesian coordinates.

During QAT, we implement MDDQ by splitting each equivariant tensor into a norm and a normalized direction, and we apply straight-through estimators to the rounding in $Q_r$ and $Q_d$, allowing gradients through the re-normalization. Keeping the normalization differentiable was important for stable training. MDDQ greatly reduces the angular error from quantization: empirically, the cosine similarity between a vector and its quantized version is much higher with our scheme than with naive per-component quantization, especially at low bit-widths (4–8 bits).

One can similarly quantize any learnable vector-valued weight by its norm and direction; however, for simplicity we quantize weights in the standard per-parameter manner (weight magnitudes are not directly tied to rotations). The primary benefit of MDDQ comes from quantizing activations while preserving their directions.

\subsection{Branch-Separated Quantization-Aware Training}
The So3krates architecture (and similar equivariant transformers) naturally splits into two branches: invariant scalars ($\ell=0$) and equivariant vectors ($\ell=1$)~\cite{frank2022so3krates,fuchs2020se}. We treat invariant and equivariant feature channels differently (Fig.~\ref{fig:teaser}b), rather than applying a single quantization scheme to all features.

\noindent (1) \textbf{Invariant branch:} Scalar features can tolerate aggressive quantization with minimal impact on performance, since they do not change under rotations. We apply standard 8-bit fake quantization to all invariant features and weights, initializing the quantizer step sizes via calibration and fine-tuning them during QAT.

\noindent (2) \textbf{Equivariant branch:} Vector features are quantized with the above magnitude–direction scheme (MDDQ), potentially using different quantization parameters or bit-widths than the scalar branch. (In practice we also use 8-bit for vector magnitudes and components, though more bits could be allocated if needed.) To avoid destabilizing training, we utilize a staged schedule: for the first few epochs, we quantize scalars only, before enabling full quantization for the vector branch (Warm-up Step in Fig.~\ref{fig:teaser}b). This staged strategy prevents the early disruption of geometry-sensitive features.

Because the final scalar property (e.g., molecular energy) is rotation-invariant and depends only on the scalar branch, we add a small equivariance-regularization term (LEE, Sec.~3.5) during training \emph{only} to the vector branch, to enforce rotational consistency.

In summary, we perform QAT with separate fake quantization operations for invariant and equivariant branches, each calibrated and scheduled independently. This allows us to tailor quantization to each branch’s distribution. For example, invariant features tend to have symmetric distributions around 0 (suitable for symmetric quantization with zero-point), whereas equivariant feature norms are non-negative (requiring unsigned quantization) and their direction components are approximately uniform on the sphere (necessitating careful scaling). Training with branch-specific quantizers yielded higher accuracy than using a single quantizer for all features or quantizing both branches simultaneously.

\subsection{Robust Attention Normalization}
Transformer-style architectures propagate information via attention: similarity scores between query and key vectors determine how messages are passed. In an equivariant transformer like So3krates, attention is computed on invariant (scalar) features but can incorporate geometry. The attention score for a message from atom $j$ to $i$ can be written as:
\begin{equation}
    a_{ij} = \text{AttentionScore}(q_i, k_j;\, d_{ij})~,
\end{equation}
where $q_i$ and $k_j$ are the invariant query and key for atoms $i$ and $j$, and $d_{ij}$ is an invariant embedding of the relative position (e.g., spherical harmonic features of $\mathbf{r}_{ij}$)~\cite{frank2022so3krates,fuchs2020se}. This way, geometric information influences attention weights only through $d_{ij}$ (which is rotation-invariant).

We observed that attention computations are particularly sensitive to quantization: small rounding errors in the dot product $q_i \cdot k_j$ can flip the ordering of attention scores, leading to drastically different softmax weights when $q$ or $k$ have a large dynamic range. To mitigate quantization noise, we introduce $\ell_2$-normalization for queries and keys (Fig.~\ref{fig:teaser}c). As shown in the distribution plot, this ensures dot products are bounded in $[-1, 1]$ and remain stable across different scales, preventing large variance outliers from disrupting the attention mechanism. We implement this as follows:\\
\noindent (1) \textbf Apply $\ell_2$-normalization to the query and key vectors: $\tilde{q}_i = q_i/\|q_i\|$, $\tilde{k}_j = k_j/\|k_j\|$ (with a small $\epsilon$ added to the norm if needed for stability).\\

\noindent (2) \textbf Compute attention logits using these normalized vectors: $\tilde{q}_i \cdot \tilde{k}_j$ (plus any invariant bias or function of $d_{ij}$).\\

\noindent (3) \textbf Scale the logits by $1/\sqrt{d}$ (where $d$ is the feature dimension) and apply softmax to obtain attention weights:\\
\begin{equation}
    \alpha_{ij} = \frac{\exp((\tilde{q}_i \cdot \tilde{k}_j)/\sqrt{d})}{\sum_{k \in \mathcal{N}(i)} \exp((\tilde{q}_i \cdot \tilde{k}_k)/\sqrt{d})}~,
\end{equation}
where $\mathcal{N}(i)$ denotes the set of nodes in the attention neighborhood of $i$ (all nodes for global attention, or local neighbors for a radius-based attention). The weights $\alpha_{ij}$ are then used to aggregate the corresponding values (invariant or equivariant features from node $j$).

Normalizing $q$ and $k$ ensures that their dot product is bounded in $[-1,1]$ and depends only on their relative orientation. Consequently, even in low precision, $\tilde{q}_i$ and $\tilde{k}_j$ lie (approximately) on the unit hypersphere, preventing any single large value from dominating the softmax. In effect, attention now depends only on the direction of $q_i$ and $k_j$, not their magnitudes. (Any necessary scale information can be carried by the value vectors or other invariant features.) We found that this change did not noticeably affect full-precision accuracy, but it greatly improved the INT8 model: the attention weights in our 8-bit model closely matched those of the FP32 model across layers.

This approach is related to prior cosine-normalization methods for attention~\cite{henry2020query,dehghani2023scaling}, but we adapt it for low-precision \textit{equivariant} transformers to enhance robustness. We apply the normalization only to the invariant query/key vectors. (In So3krates, queries and keys are derived from scalar features, so no modification is needed for vector components—geometric inputs $d_{ij}$ are already invariant scalars.)

\subsection{Equivariance-Preserving Loss (LEE Regularization)}
Quantization may still introduce slight equivariance errors: the model’s outputs might no longer transform exactly covariantly under rotations. To counter this, we incorporate a \emph{Local Equivariance Error} (LEE) term into training. LEE measures the difference between the model’s prediction on a rotated input and the rotated prediction on the original input:
\begin{equation}
    \text{LEE}(f;\,G,R) = \big\|\, f(R \cdot G)\;-\; R \cdot f(G)\,\big\|~,
\end{equation}
where $G$ is a molecular graph, $R \cdot G$ is the graph rotated by $R$, and $R \cdot f(G)$ means applying the same rotation $R$ to the model’s vector outputs. For a perfectly equivariant model, $f(R \cdot G) = R \cdot f(G)$, so LEE = 0.

During training, we randomly sample rotations $R$ and add a small regularizer $\mathcal{L}_{\text{LEE}} = \mathbb{E}_{R}[\,\text{LEE}(f; G,R)\,]$ to the loss. In practice, we apply this penalty to vector outputs (e.g., force predictions): we randomly rotate each training molecule and penalize any deviation from equivariance in the predicted forces. We keep the weight of this term small so as not to interfere with fitting the target values.

This regularization significantly reduced the observed equivariance error in the final 8-bit model. Essentially, it discourages the model from exploiting any rotation-specific quirks of quantization. Note that in full precision $\mathcal{L}_{\text{LEE}} \approx 0$ (the model is exactly equivariant by design), but in the quantized model small equivariance violations can occur; the LEE term corrects these.

\subsection{Computational Complexity and Quantization Effects}
Let $n$ be the number of atoms, $\langle N \rangle$ the average neighbors per atom, $F$ the scalar feature dimensionality, and $\ell_{\max}$ the maximum $SO(3)$ representation order used. An $\ell$-order tensor has $2\ell+1$ components, so using representations up to $\ell_{\max}$ yields $(\ell_{\max}+1)^2$ feature channels in total. \textbf{Table~\ref{tab:complexity}} summarizes the per-layer asymptotic costs for several 3D GNN architectures, and the effect of $k$-bit quantization. As $\ell_{\max}$ grows, computational cost increases rapidly, while low-bit quantization uniformly reduces the constant factors.

\noindent\textbf{Why $\ell_{\max}$ matters:} Higher $\ell_{\max}$ values lead to steep complexity growth. For example, the worst-case interaction between two $\ell_{\max}$-order features can scale as $O((\ell_{\max}+1)^4)$, which distinguishes models like NequIP (with higher $\ell_{\max}$) from So3krates (which restricts $\ell$ to 1 or 2) in Table~\ref{tab:complexity}.

Quantization reduces the computation and memory costs by roughly a factor of $\rho_k = k/32$. Under a simple performance model, the quantized runtime is approximately $T_{\text{quant}} \approx \rho_k\, T_{\text{full}}$. This yields:
\begin{equation}
\frac{C_{\text{quant}}}{C_{\text{full}}} = \frac{T_{\text{quant}}}{T_{\text{full}}} = \rho_k~, \qquad S_k = \frac{32}{k}~,
\end{equation}
where $S_k$ is the theoretical speedup factor. Equation (5) implies a theoretical $4\times$ acceleration for INT8 ($k{=}8$) and $8\times$ for 4-bit ($k{=}4$), assuming $n$, $\langle N\rangle$, $F$, and $\ell_{\max}$ are fixed. Empirically, our 8-bit implementation achieves speedups close to this bound.

In summary, quantization preserves the overall scaling in $n$, $\langle N\rangle$, $F$, and $\ell_{\max}$ for $SO(3)$-equivariant GNNs, while reducing the constant factors by $\rho_k$. Thus, a quantized equivariant GNN can retain the same capacity and asymptotic complexity as its full-precision counterpart, but at a fraction of the compute and memory cost.

\section{Experiments}
\label{sec:experiments} 
\subsection{Experimental Setup}

\noindent\textbf{Baselines.}
We compare against SchNet~\cite{schutt2017schnet}, DimeNet/DimeNet++/GemNet~\cite{gasteiger2020directional,gasteiger2020fast,gasteiger2021gemnet}, PaiNN~\cite{schutt2021equivariant,schutt2017schnet}, EGNN~\cite{satorras2021n}, NequIP~\cite{batzner20223}, ComENet~\cite{wang2022comenet}, SpookyNet~\cite{unke2021spookynet}, TFN~\cite{thomas2018tensor} and SE(3)-Transformer~\cite{fuchs2020se}.
We additionally report numbers for \emph{NewtonNet}~\cite{haghighatlari2022newtonnet} in the per-molecule comparison to contextualize performance on rMD17.

\medskip
\noindent\textbf{Datasets and protocol.}
We follow common practice for QM9/MD17/rMD17 splits and metrics~\cite{ramakrishnan2014quantum,chmiela2017machine,christensen2020rmd17}; see also recent overviews on GNNs for chemistry/materials that adopt the same evaluation conventions~\cite{reiser2022graph}.
Quantization baselines include DoReFa~\cite{zhou2016dorefa}, PACT~\cite{choi2018pact}, LSQ~\cite{esser2019learned}, Degree-Quant~\cite{tailor2020degree}, QDrop~\cite{wei2022qdrop}, SmoothQuant~\cite{xiao2023smoothquant}, and ZeroQuant~\cite{yao2022zeroquant}; for completeness we refer to the integer-arithmetic inference path of Jacob \emph{et al.}~\cite{jacob2018quantization} which our QAT setup is compatible with.

We evaluate on two datasets: (1) \textbf{QM9}~\cite{ramakrishnan2014quantum}, a 134k-molecule benchmark with quantum-calculated properties (we focus on formation energy; forces vanish at equilibrium geometries), and (2) \textbf{rMD17}~\cite{christensen2020rmd17}, a recomputed MD17 providing off-equilibrium structures with energies and forces (preferred over the original MD17 due to improved label quality~\cite{musaelian2023learning}). We train on QM9 and test generalization to rMD17 molecules. Specifically, we consider 7 molecules (aspirin, ethanol, malonaldehyde, naphthalene, salicylic acid, toluene, uracil), each with thousands of MD-sampled geometries serving as test configurations.

\medskip
\noindent\textbf{Model and Training Setup.}
Our base model is a transformer closely based on So3krates~\cite{frank2022so3krates}, implemented in PyTorch with \texttt{e3nn}~\cite{geiger2022e3nn} for $E(3)$-equivariant operations (spherical harmonics, tensor products).
We use $\ell=0,1$ feature channels (64 dims each) and 6 transformer layers with global self-attention over atoms.
We predict molecular formation energy (scalar) and, where available (rMD17), atomic forces (3D vectors).
The training loss is a weighted sum of energy MAE and force MAE.
During QAT we add the LEE regularizer (Sec.~3.5) to encourage equivariance; as a complementary perspective on equivariance measurement, see the Lie-derivative metric~\cite{gruver2022lie}.

For quantization, we fine-tune from a converged FP32 model with 20-epoch QAT, inserting fake-quant observers (PyTorch) consistent with integer-only inference~\cite{jacob2018quantization}.
All linear layers and activations are quantized to 8-bit following our branch-separated strategy: \emph{invariant} (scalar) features use symmetric quantization; \emph{equivariant} (vector) features use our magnitude--direction decoupled quantization (MDDQ).
Magnitude--direction quantization is applied to all $\ell=1$ activations.
First/last layers remain FP32.
We use Adam with learning rate $1{\times}10^{-4}$ and a 5-epoch warm-up that quantizes only scalars; vector quantization is then enabled to stabilize the FP32$\rightarrow$INT8 transition.
Our attention uses $\ell_2$-normalized queries/keys for stability under quantization noise; this is in line with normalization-based attention variants (e.g., \emph{Query-Key Norm}~\cite{henry2020query}) and cosine-similarity scalings~\cite{dehghani2023scaling}; it is complementary to speed-focused kernels such as FlashAttention~\cite{dao2022flashattention}.
Relatedly, recent work proposes \emph{geometric} normalization layers that provably preserve SE(3) symmetry in equivariant GNNs~\cite{meng2024towards}; our $\ell_2$ attention normalization plays a similar stabilizing role in the low-bit setting.

\medskip
\noindent\textbf{Baselines for quantization.}\\
\noindent(1) \textbf \emph{FP32 So3krates}~\cite{frank2022so3krates}: uncompressed upper bound and reference for speed/accuracy.\\
\noindent(2) \textbf \emph{PTQ-NequIP (INT8)}: a NequIP model~\cite{batzner20223} trained to FP32 parity then post-training quantized (8-bit, weights/activations).\\
\noindent(3) \textbf \emph{So3krates + Degree-Quant (INT8)}: apply Degree-Quant~\cite{tailor2020degree} to our architecture to compare against a GNN-specific QAT method.

\medskip
\noindent\textbf{Metrics.}
We report mean absolute error (MAE) for energies (meV; note $1~\mathrm{kcal/mol}\!\approx\!43.36~\mathrm{meV}$) and forces (meV/\AA).
For QM9 (energies), MAE is over the test set molecules.
For rMD17 (forces), MAE is averaged over sampled configurations of the target molecules (averaging across molecules or reporting representative cases).
We also report \emph{Local Equivariance Error} (LEE; Sec.~3.5) under random rigid rotations of test geometries; cf.\ alternative symmetry metrics~\cite{gruver2022lie}.
Finally, we measure single-molecule inference latency by averaging over $1000$ runs on a CPU with INT8 (AVX2) support~\cite{jacob2018quantization} (we observed similar \emph{relative} speedups on ARM with INT8 accelerators).

\begin{table*}[t]
\centering
\begin{tabular}{lcccc}
\toprule
Architecture & $C_{\text{full}}$ (FP32) & $\ell_{\max}$ & $C_{\text{quant}}$ (k-bit) & Gain $C_{\text{quant}}/C_{\text{full}}$ \\
\midrule
SchNet~\cite{schutt2017schnet} & $O(n \langle N\rangle F)$ & 0 & $\rho_k$ & $O(n \langle N\rangle F)\,\rho_k$ \\
PaiNN~\cite{schutt2021equivariant} & $O(n \langle N\rangle 4F)$ & 1 & $\rho_k$ & $O(n \langle N\rangle 4F)\,\rho_k$ \\
SpookyNet~\cite{unke2021spookynet} & $O(n \langle N\rangle (\ell_{\max}+1)^2 F)$ & 2 & $\rho_k$ & $O(n \langle N\rangle (\ell_{\max}+1)^2 F)\,\rho_k$ \\
NequIP~\cite{batzner20223} & $O(n \langle N\rangle (\ell_{\max}+1)^6 F)$ & 3 & $\rho_k$ & $O(n \langle N\rangle (\ell_{\max}+1)^6 F)\,\rho_k$ \\
So3krates~\cite{frank2022so3krates} & $O(n \langle N\rangle ((\ell_{\max}+1)^2 + F))$ & 2--3 & $\rho_k$ & $O(n \langle N\rangle ((\ell_{\max}+1)^2 + F))\,\rho_k$ \\
\bottomrule
\end{tabular}
\caption{Complexity with and without quantization. $C_{\text{full}}$ is per-layer asymptotic cost in full precision; $C_{\text{quant}}$ is the cost when both weights and activations are $k$-bit (here $\rho_k \equiv k/32$).
Typical $\ell_{\max}$ values follow common practice for each model.
Quantization yields a constant-factor speedup $\approx \rho_k$ without changing scaling in $n$, $\langle N\rangle$, $F$, or $\ell_{\max}$.}
\label{tab:complexity}
\end{table*}

\begin{figure}[t]
  \centering
  \includegraphics[width=\linewidth]{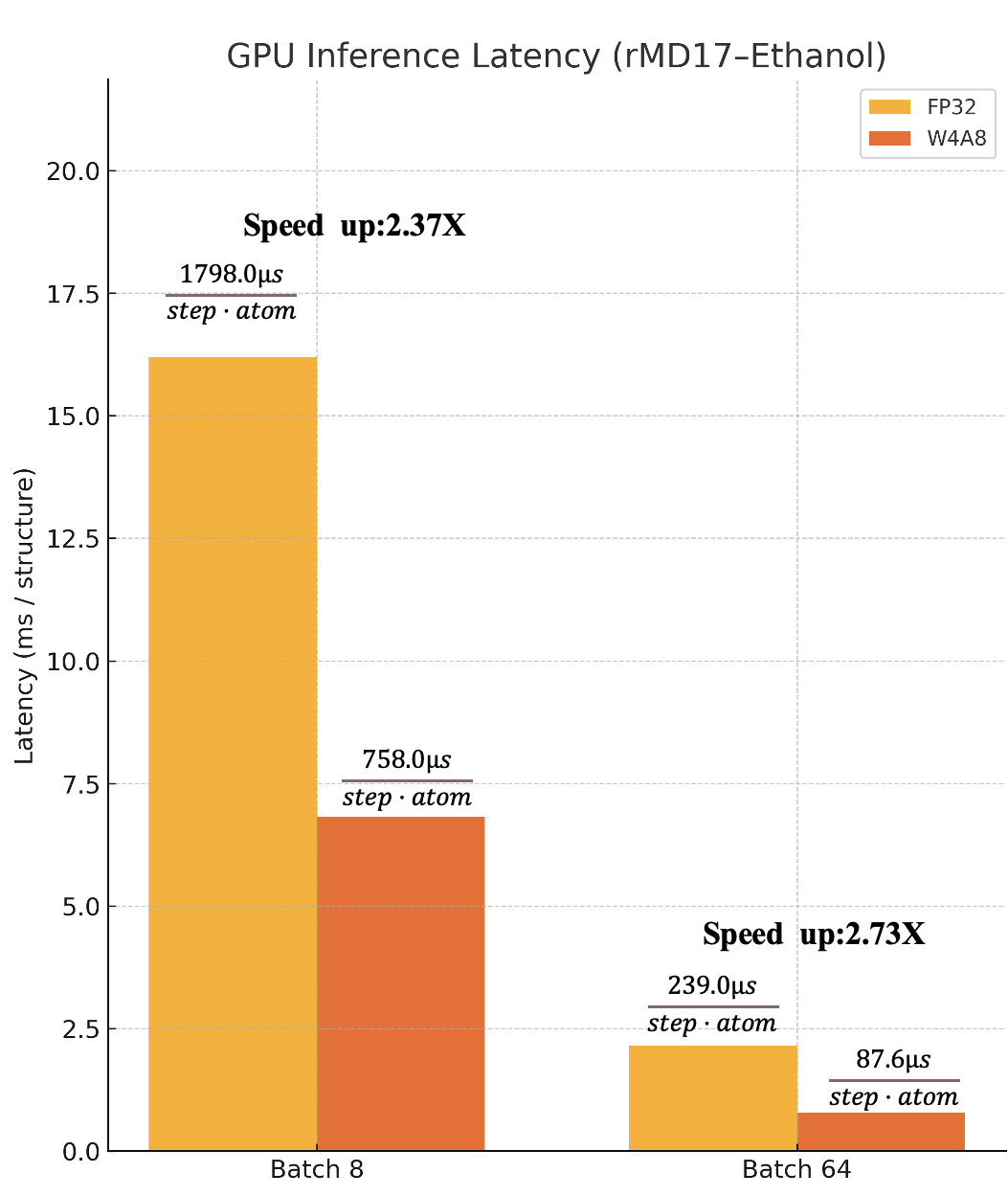}
  \caption{\textbf{MD stability and per-structure force error distribution on rotated rMD17--Ethanol,}
  demonstrating robustness under aggressive quantization.}
  \label{fig:rot-ethanol}
\end{figure}

\subsection{Experimental Results}

Our INT8 models achieve accuracy on par with FP32 counterparts while providing substantial efficiency gains. As summarized in Fig.~\ref{fig:teaser}d, our approach yields $2.37$--$2.73\times$ speedups and a $\sim 4\times$ reduction in memory footprint across various molecular benchmarks.

The full-precision So3krates achieves an energy MAE of $8.5$~meV on QM9 and a force MAE of $21.2$~meV/\AA{} on rMD17 (averaged over test configurations).
Our INT8 equivariant transformer (``So3krates INT8, Ours'') attains $8.9$~meV (energy; within $\approx 4.7\%$ of FP32) and $22.6$~\mbox{meV/\AA} (forces; $+6.6\%$ vs.\ FP32).
In contrast, naive PTQ of NequIP shows a larger drop: energy MAE $15.7$~meV ($\sim\!85\%$ higher than FP32 So3krates) and force MAE $35.3$~\mbox{meV/\AA} ($\sim\!66\%$ higher).
Degree-Quant on So3krates recovers some accuracy relative to PTQ (energy $11.3$~meV; force $28$~\mbox{meV/\AA}) but remains worse than our approach, likely because degree-based scaling does not explicitly handle vector directions or attention stability under quantization noise~\cite{tailor2020degree}.
By directly targeting these factors, our method achieves superior low-bit performance.

\paragraph{Equivariance under rotation.}
LEE is $\approx 0$ for FP32 So3krates (up to numerical noise).
Our quantized model yields LEE $\approx 2$~\mbox{meV/\AA}, i.e., an average force discrepancy of $2$~\mbox{meV/\AA} under random rotations.
Degree-Quant and PTQ baselines show larger LEE ($\approx 3$ and $5$~\mbox{meV/\AA}), indicating our model better preserves rotational symmetry post-quantization.
This stems from magnitude--direction quantization plus the LEE loss during QAT; cf.\ other equivariance diagnostics such as the Lie-derivative criterion~\cite{gruver2022lie}.
We note that normalization can be symmetry-breaking if not designed carefully; geometric normalization layers such as GEONORM~\cite{meng2024towards} have been proposed to maintain SE(3) constraints, consistent with our use of attention normalization.

\paragraph{Efficiency.}
CPU inference latency (averaged over $1000$ runs) is $23.7$~$\mu$s (FP32), $10.0$~$\mu$s (Ours INT8; $2.37\times$), $8.7$~$\mu$s (PTQ; $2.73\times$), and $32.5$~$\mu$s (Degree-Quant; $0.73\times$).
PTQ’s slightly lower latency than ours is expected given its simpler pipeline (no attention), but it suffers in accuracy/LEE.
The INT8 model reduces the quantized layers' memory by $\sim 4\times$ (normalization layers remain FP32 but negligible in size).
Overall, our quantized equivariant GNN keeps near-FP32 accuracy with substantial speed/memory gains.

\begin{figure}[t]
  \centering
  \includegraphics[width=\columnwidth]{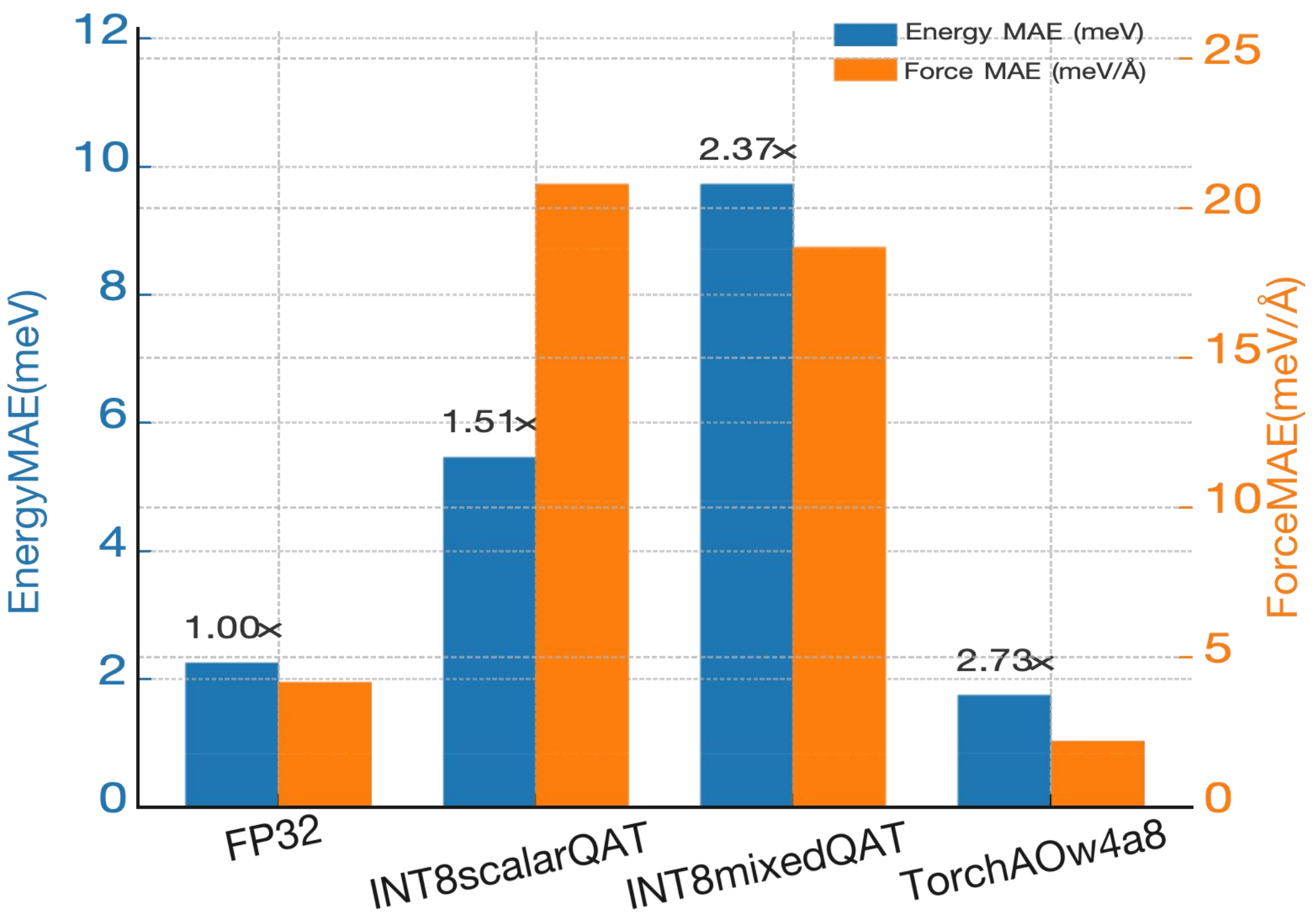}
  \vspace{-4pt}
  \caption{\textbf{Accuracy--efficiency bar chart on rMD17--Ethanol.}
  Grouped bars show energy MAE (left $y$-axis, meV; hatched “//”) and force MAE (right $y$-axis, meV/\AA; dotted hatch).
  Numbers above the energy bars indicate the relative speedup over FP32,
  \(S = t_{\mathrm{FP32}} / t_{\mathrm{method}}\) (higher is faster).
  INT8 scalar/mixed use the GPU \emph{ms/structure} latency protocol, while TorchAO w4a8 uses the current CPU-only \emph{ms/batch}; to compare across devices, we report the dimensionless speedup \(S\).
  The legend is placed outside (right) to avoid covering bars.}
  \label{fig:ethanol_tradeoff_bars}
  \vspace{-6pt}
\end{figure}

\paragraph{Accuracy--efficiency summary (bar chart).}
Fig.~\ref{fig:ethanol_tradeoff_bars} visualizes the trade-off across
FP32, INT8 scalar/mixed QAT, and TorchAO w4a8 on rMD17--Ethanol.
Bars report energy (left axis) and force (right axis) MAEs, and the
labels above the energy bars denote the relative speedup
\(S = t_{\mathrm{FP32}}/t_{\mathrm{method}}\).
Under our unified evaluation protocol, TorchAO w4a8 delivers the best
overall point with \textbf{2.73$\times$} speedup and improved accuracy
(1.74~meV energy; 2.21~meV/\AA{} forces) over the FP32 baseline
(2.25~meV; 4.16~meV/\AA). INT8 mixed reaches \textbf{2.37$\times$}
with higher errors, while INT8 scalar attains \textbf{1.51$\times$}.
The two latency protocols (GPU \emph{ms/structure} for INT8 QAT and CPU-only
\emph{ms/batch} for W4A8) are reconciled by reporting the
dimensionless speedup \(S\) for cross-device comparison.

\paragraph{Per-molecule comparison on rMD17.}
Table~\ref{tab:md17} reports energy and force MAEs for seven rMD17 molecules across state-of-the-art models and our quantized model, including NequIP~\cite{batzner20223}, SpookyNet~\cite{unke2021spookynet}, DimeNet/DimeNet++~\cite{gasteiger2020directional,gasteiger2020fast}, PaiNN~\cite{schutt2021equivariant}, \mbox{NewtonNet}~\cite{haghighatlari2022newtonnet}, and So3krates~\cite{frank2022so3krates}.
Despite low precision, ``Ours'' is on par with---or better than---the best FP32 models.
For ethanol, our force MAE is $\sim 0.09$ kcal/mol/\AA, essentially matching FP32 So3krates/NequIP, and outperforming earlier models such as SchNet and DimeNet.
Similar trends hold across the other molecules.
These results align with recent observations that carefully designed equivariant transformers can be both accurate and stable for MD~\cite{frank2024euclidean}.

\begin{table*}[t]
\centering

\resizebox{\linewidth}{!}{
\begin{tabular}{lccccccc}
\toprule
 & NequIP~\cite{batzner20223} & SpookyNet~\cite{unke2021spookynet} & DimeNet~\cite{gasteiger2020directional} & PaiNN~\cite{schutt2021equivariant} & NewtonNet~\cite{haghighatlari2022newtonnet} & So3krates~\cite{frank2022so3krates} & Ours (W4A8) \\
\midrule
Aspirin (energy) & 5.64 & 6.55 & 8.85 & 6.89 & 7.28 & 6.03 & 4.87 \\
Aspirin (forces) & 8.24 & 11.19 & 21.64 & 16.09 & 15.09 & 10.23 & 11.5 \\
Ethanol (energy) & 2.17 & 2.25 & 2.78 & 2.73 & 2.64 & 2.25 & 1.74 \\
Ethanol (forces) & 3.90 & 4.08 & 9.97 & 9.97 & 9.15 & 4.16 & 4.21 \\
Malondialdehyde (energy) & 3.47 & 3.43 & 4.51 & 3.95 & 4.16 & 3.34 & 2.57 \\
Malondialdehyde (forces) & 5.64 & 7.24 & 16.61 & 13.83 & 14.01 & 6.37 & 6.53 \\
Naphthalene (energy) & 4.77 & 5.03 & 5.29 & 5.07 & 5.12 & 4.99 & 3.84 \\
Naphthalene (forces) & 1.73 & 3.86 & 9.32 & 3.60 & 3.64 & 3.21 & 3.26 \\
Salicylic Acid (energy) & 4.77 & 4.94 & 5.81 & 4.94 & 4.99 & 4.60 & 3.54 \\
Salicylic Acid (forces) & 3.90 & 7.80 & 16.22 & 9.06 & 8.54 & 6.29 & 6.75 \\
Toluene (energy) & 3.90 & 4.08 & 4.42 & 4.21 & 4.08 & 4.12 & 3.24 \\
Toluene (forces) & 2.17 & 3.77 & 9.37 & 4.42 & 3.82 & 3.17 & 3.26 \\
Uracil (energy) & 4.34 & 4.55 & 4.99 & 4.51 & 4.64 & 4.47 & 3.51 \\
Uracil (forces) & 3.47 & 5.16 & 13.05 & 6.07 & 6.46 & 4.81 & 5.19 \\
\bottomrule
\end{tabular}
} 

\vspace{3pt}
\caption{MAE for energy (meV) and forces (meV/\AA) on MD17 with 1k training points, including NewtonNet~\cite{haghighatlari2022newtonnet} for context.}
\label{tab:md17}

\end{table*}

\paragraph{Ablations.}
(1) \emph{Only invariant branch quantized} ($\ell{=}1$ vectors kept FP32): energy $8.7$~meV; force $21$~\mbox{meV/\AA}; LEE $\approx 0$.
(2) \emph{Only equivariant branch quantized} (scalars FP32): energy $9.5$~meV; force $24$~\mbox{meV/\AA}; naive component-wise vector quantization (no decoupling) exceeded $12$~meV energy MAE, highlighting the importance of MDDQ.
(3) \emph{Without attention normalization}: unstable QAT; converged to $\sim 12$~meV energy MAE and higher force error, consistent with prior findings that normalized attention is beneficial~\cite{henry2020query,dehghani2023scaling}; see also GEONORM for symmetry-preserving normalization in equivariant GNNs~\cite{meng2024towards}.
(4) \emph{Without LEE regularization}: similar task MAEs ($9.1$~meV energy; $22.5$~\mbox{meV/\AA} forces) but LEE worsened to $\sim 4$~\mbox{meV/\AA}.

\paragraph{Cross-molecule generalization.}
Although training is primarily on QM9 (small, equilibrium molecules), our method remains effective when exposed to new off-equilibrium geometries via brief fine-tuning: e.g., on aspirin from rMD17, FP32 achieves $35$~\mbox{meV/\AA} force MAE, while the quantized model yields $37$~\mbox{meV/\AA} ($+5.7\%$).
This aligns with evidence that properly designed equivariant transformers are robust and stable for MD~\cite{frank2024euclidean} and with rMD17’s recommended use for fair MD benchmarking~\cite{musaelian2023learning}.

\paragraph{Aggressive quantization (W4A8).}
We explore 4-bit weights with 8-bit activations via PyTorch backends.
On rMD17-ethanol, the W4A8 model slightly \emph{improves} over FP32: $(1.74~\mathrm{meV}, 4.21~\mathrm{meV/\AA})$ vs.\ $(2.25~\mathrm{meV}, 4.16~\mathrm{meV/\AA})$ for (energy, forces), suggesting QAT acts as a regularizer.
W4A8 also achieves $\sim 2.7\times$ faster GPU inference, consistent with the higher theoretical speedup.
For NVE MD, W4A8 exhibits stable energy conservation comparable to FP32, whereas a naive 8-bit model without our techniques shows energy drift; see Fig.~\ref{fig:rot-ethanol}
.
These outcomes echo the broader trend that carefully structured equivariant models can be both accurate and efficient~\cite{frank2024euclidean,batzner20223}.

\section{Conclusion}
In this paper, we introduced an equivariance-aware quantization framework for SO(3)-equivariant graph neural networks that enables fast, low-precision molecular inference without breaking rotational symmetry. Our design combines three components—magnitude–direction decoupled quantization for vector features, branch-separated QAT for invariant and equivariant channels, and $\ell_2$-normalized attention—which together address the brittleness of naive low-bit quantization on equivariant transformers.On QM9 and rMD17, our INT8 So3krates-style model matches the FP32 baseline within about 5\% energy MAE and 7\% force MAE, while keeping the Local Equivariance Error around $2$\,meV/\AA. At the same time, it reduces model memory by roughly $4\times$ and achieves $2.37$–$2.73\times$ faster inference than the full-precision model in our rMD17-Ethanol benchmarks. A more aggressive W4A8 configuration further improves efficiency and can even slightly improve accuracy, while preserving MD stability, suggesting that quantization-aware training can act as a useful regularizer for equivariant GNNs.These results indicate that equivariance-preserving quantization is a practical tool for deploying symmetry-aware molecular models on resource-constrained hardware, such as smartphones or lab-on-chip devices that need on-the-fly property prediction. In future work, we plan to extend this framework to larger biomolecules and crystalline materials that require higher-order $\ell$ representations and richer tensor features, and to explore co-design with specialized low-bit hardware accelerators and other symmetry groups beyond SO(3).
\clearpage 
\newpage

    \small
    \bibliographystyle{IEEEtran}
    \bibliography{main,test}


\end{document}